\documentclass{article}
\usepackage{ijcai19}

\usepackage{times}
\usepackage{soul}
\usepackage{url}
\usepackage[hidelinks]{hyperref}
\usepackage[utf8]{inputenc}
\usepackage[small]{caption}
\usepackage{graphicx}
\usepackage{amsmath}
\usepackage{booktabs}
\usepackage{csquotes}
\graphicspath{ {./images/} }

\title{Curriculum Learning Strategies for Hindi-English Codemixed Sentiment Analysis\footnote{This work was presented at 2nd Workshop on Humanizing AI (HAI) at IJCAI'19 in Macao, China.}}

\author{
Anirudh Dahiya$^1$\footnote{Contact Author}\and
Neeraj Battan$^2$\and
Manish Shrivastava$^1$\And
Dipti Mishra Sharma$^1$\\
\affiliations
$^1$LTRC, IIIT Hyderabad, India\\
$^2$CVIT, IIIT Hyderabad, India\\
\emails
\{anirudh.dahiya, neeraj.battan\}@research.iiit.ac.in,
dipti@iiit.ac.in,
m.shrivastava@iiit.ac.in
}

\begin{document}

\maketitle

\begin{abstract}
Sentiment Analysis and other semantic tasks are commonly used for social media textual analysis to gauge public opinion and make sense from the noise on social media. The language used on social media not only commonly diverges from the formal language, but is compounded by codemixing between languages, especially in large multilingual societies like India.

Traditional methods for learning semantic NLP tasks have long relied on end to end task specific training, requiring expensive data creation process, even more so for deep learning methods. This challenge is even more severe for resource scarce texts like codemixed language pairs, with lack of well learnt representations as model priors, and task specific datasets can be few and small in quantities to efficiently exploit recent deep learning approaches. To address above challenges, we introduce curriculum learning strategies for semantic tasks in code-mixed Hindi-English (Hi-En) texts, and investigate various training strategies for enhancing model performance. Our method  outperforms the state of the art methods for Hi-En codemixed sentiment analysis by 3.31\% accuracy, and also shows better model robustness in terms of convergence, and variance in test performance.
\end{abstract}

\section{Introduction}

Codemixing is the phenomenon of intermixing linguistic units from two or more languages in a single utterance, and is especially widespread in multilingual societies across the world \cite{muysken2000bilingual}. With increasing internet access to such large populations of multilingual speakers, there is active ongoing research on processing codemixed texts on online socialmedia communities such as Twitter and Facebook \cite{singh-etal-2018-twitter,DBLP:AmeyaPrabhuJSV16}. Not only do these texts contain a diverse variety of language use spanning the formal and colloquial spectra, such texts also pose challenging problems such as out of vocabulary words, slangs, grammatical switching and structural inconsistencies.

Previously, various approaches \cite{DBLP:AmeyaPrabhuJSV16,DBLP:madanEmsemble,singh-etal-2018-twitter,singh2018twitter} have focused on task specific datasets and learning architectures for syntactic and semantic processing for codemixed texts. This has facilitated developement of various syntactic and semantic task specific datasets and neural architectures, but has been limited by the expensive efforts towards annotation. As a result, while these efforts have enabled processing of codemixed texts, they still suffer from data scarcity and poor representation learning, and the small individual dataset sizes usually limiting the model performance.

Curriculum Learning, as introduced by \cite{bengio2009curriculum} is \enquote{to start small, learn easier aspects of the task or easier subtasks, and then gradually increase the difficulty level}. They also draw parallels with human learning curriculum and education system, where different concepts are introduced in an order at different times, and has led to advancement in research towards animal training \cite{krueger2009flexible}. Previous experiments with tasks like language modelling \cite{bengio2009curriculum}, Dependency Parsing, and entailment \cite{DBLP:journals/corr/HashimotoXTS16} have shown faster convergence and performance gains by following a curriculum training regimen in the order of increasingly complicated syntactic and semantic tasks. \cite{DBLP:journals/corr/abs-1802-03796} also find theoretical and experimental evidence for curriculum learning by pretraining on another task leading to faster convergence.

With this purview, we propose a syntactico-semantic curriculum training strategy for Hi-En codemixed twitter sentiment analysis. We explore various pretraining strategies encompassing Language Identification, Part of Speech Tagging, and Language Modelling in different configurations. We investigate the role of different transfer learning strategies by changing learning rates and gradient freezing to prevent catastrophic forgetting and interference between source and target tasks. We also propose a new model for codemixed sentiment analysis based on character trigram sequences and pooling over time for representation learning. We investigate the convergence rate and model performance across various learning strategies, and find faster model convergence and performance gains on the test set.

\section{Related Work}
Research on semantic and syntactic processing of codemixed texts has increasingly gained attention, and various approaches have been proposed to this end. \cite{DBLP:AmeyaPrabhuJSV16} released a dataset comprising user comments on Facebook pages, and proposed a convolutions over character embeddings approach towards sentiment analysis for Hi-En Codemixed texts. \cite{DBLP:madanEmsemble} propose a character trigram approach coupled with an ensemble of an RNN and a Naive Bayes classifier towards sentiment analysis for codemixed data.

More generally for monolingual sentiment analysis, RNNs and other sequential deep learning models have shown to be successful. \cite{socher-etal-2012-semantic} obtained significant performance improvement by incorporating compositional vector representations over single vector representations. \cite{zheng2018leftcenterright} take a different approach by capturing the most important words on either side to perform targeted sentiment analysis. Their LSTM based model uses context2target attention to achieve better benchmark performance on three datasets.  \cite{singh2018twitter} developed a dataset for Hi-En codemixed Part of Speech tagging, and proposed a CRF based approach. \cite{singh2018twitter} developed a dataset for Hindi English Codemixed Language Identification and NER, and propose a CRF based approach with handcrafted features for Named Entity Recognition. 

Bengio et. al.\shortcite{bengio2009curriculum} introduced curriculum learning approaches towards both vision and language related task, and show significant convergence and performance gains for language modelling task. \cite{DBLP:journals/corr/HashimotoXTS16} propose a hierarchical multitask neural architecture with the lower layers performing syntactic tasks, and the higher layers performing the more involved semantic tasks while using the lower layer predictions. \cite{DBLP:journals/corr/abs-1808-10485} also propose a syntactico semantic curriculum with chunking, semantic role labelling and coreference resolution, and show performance gains over strong baselines. Like \cite{DBLP:journals/corr/HashimotoXTS16}, they hypothesize the incorporation of simpler syntactic information into semantic tasks, and provide empirical evidence for the same.


\section{Datasets}

\cite{DBLP:AmeyaPrabhuJSV16} released a Hi-En codemixed dataset for Sentiment Analysis, comprising 3879 Facebook comments on public pages of Salman Khan and Narendra Modi. Comments are annotated as positive, negative and neutral based on their sentiment polarity, and are distributed across the 3 classes as 15\% negative, 50\% neutral and 35\% positive comments.

\cite{singh-etal-2018-twitter} released a twitter corpus for Part of Speech tagging for Hindi English codemixed tweets about 5 incidents, and annotated 1489 tweets with the POS tag for each token.

\cite{singh2018twitter} released a Hindi English codemixed twitter corpus for Language Identification and Named Entities, where each token is annotated for its Language as English, Hindi or rest. Apart from this, they also annotate 2763 named entities in text, classified as Person, Location or Organization.  
\begin{table}[ht]
\centering
\begin{tabular}{@{}ll@{}}
\toprule
\textbf{Symbol}      & \textbf{Description} \\ \midrule
$h_{r_i}^{(j)}$ & jth Layer's right directional hidden state at time i    \\
$h_{l_i}^{(j)}$             &  jth Layer's left directional hidden state at time i \\
$H_{i}^{(j)}$ & Left and right hidden state concatenation \\
$H_{N}^{(j)} $ & Terminal H for layer j LSTM   \\
\bottomrule
\end{tabular}
\caption{Symbol Description}
\label{table:symbol}
\end{table}

\section{Approach}

In the following subsections, we introduce our proposed model architecture for processing the above tasks in a hierarchical manner, discuss the various curriculum strategies we experiment with, and finally discuss the transfer learning techniques we explore. 

\begin{figure}[h]
\includegraphics[width=3.37in]{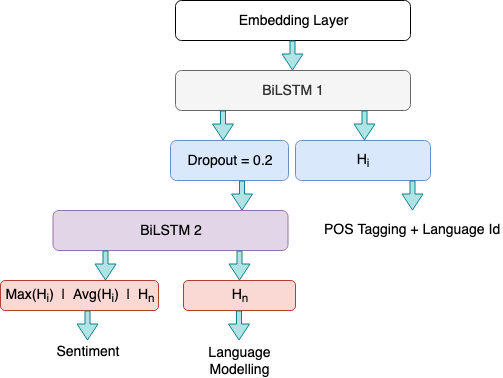}
\caption{Systematic Overview of Model Architecture}
\centering
\label{fig:model}
\end{figure}

\subsection{Model}

We case normalise the texts and mask user mentions and URLs with special characters. After tokenizing the texts, we append a token terminal \enquote{*} symbol to each token and further split each token into its constituent character trigrams. Thus, a token \enquote{girl} is split into \enquote{gir} + \enquote{l*\#}, where \enquote{\#} is the padding symbol for character trigrams. Due to the high class imbalance in Sentiment Analysis data, we perform a mixture of oversampling and undersampling between classes than simply prune the samples of the larger classes.

As shown in Figure \ref{fig:model}, the model comprises of an Embedding layer followed by two layers of bidirectional LSTMs. The embedding layer serves as a lookup table for our character trigram dense representations, and a sequence of these representations are passed on to the LSTM stack for each input sample.

Layer 1 of the LSTM stack takes the sequence of character trigram embeddings as input, and is used to predict the corresponding POS tag and the Language tag at each time step. The concatenation of the left and right directional hidden states at each time step i is passed to a standard softmax classifier to output the probability distribution over the POS tags at that timestep. Similarly, another softmax classifier outputs the language tags at each timestep.

\begin{flalign*}
&H_{i}^{(1)} =  concat([h_{r_i}^{(1)}; h_{l_i}^{(1)}])  &\\
&y_{i_{pos}} = Softmax(W_{pos}H_{i}^{(1)} + b_{pos}) &\\
&y_{i_{lang}} = Softmax(W_{lang}H_{i}^{(1)} + b_{lang}) &\\
\end{flalign*}

The LSTM Layer 2 takes the concatenated bidirectional hidden states of Layer 1 ($H_i^{(1)}$) as input to learn the sequence representation as an abstraction over the layer 1 representations. This architecture allows the semantic task to consider both the character trigram dense representation as well as "POS" and "Language" aware bidirectional representations to perform more complex semantic tasks like language modelling and sentiment analysis.

For language modelling, the concatenated terminal hidden states from the right and left directional LSTM layer 2 is passed to a standard softmax classifier, which outputs the probability distribution over the character trigram vocabulary for the next trigram in the input sequence. For sentiment analysis, we concatenate the max pooling over time, avg pooling over time, and the terminal hidden states of the Layer 2 BiLSTM to form the representation. The maxpooling and avgpooling over time representations circumvents the information loss in sequence terminal representations.  This representation is passed to a standard softmax classifier to predict the sentiment polarity over the 3 classes.

\begin{flalign*}
&H_{i}^{(2)} =  concat([h_{r_i}^{(2)}; h_{l_i}^{(2)}])  &\\
&y_{LM} = Softmax(W_{LM}H_{N}^{(2)} + b_{LM}) &\\
&H_{S}^{(2)} = concat([H_{N}^{(2)};maxpool(H_{i}^{(2)});avgpool(H_{i}^{(2)})]) &\\
&y_{sentiment} = Softmax(W_{sentiment}H_{S}^{(2)} + b_{sentiment}) &\\
\end{flalign*}

For each of the tasks described above, we train our model to optimize the cross entropy loss for the given prediction, formulated as :

\[L = -\sum_{y\in Y} y\log{p} + (1-y)log(1-p)\]

where y is the true label, and p is the predicted probability of that label by the model.

\subsection{Curriculum Training}

While our proposed model enables efficient transfer learning by progressive abstraction of representations for more complicated tasks, the highlight of the approach lies in the training regimen followed.

Curriculum learning can be seen as a sequence of training criteria \cite{bengio2009curriculum}, with increasing task or sample difficulty as the training progresses. It is also closely related with transfer learning by pretraining, especially in the case when the tasks form a logical hierarchy and contribute to the downstream tasks. With this purview, we propose a linguistic hierarchy of training tasks for codemixed languages, with further layers abstracting over the previous ones to achieve increasingly complicated tasks. Considering the codemixed nature of texts and linguistic hierarchy of information, we propose the tasks in the order of : Language Identification, Part of Speech Tagging, Language Modelling and further semantic tasks like sentiment analysis.

Since tokens in codemixed texts have distinct semantic spaces based on their source language, Language Identification can incorporate this disparity among the learnt trigram representations. Following this, the Part of Speech Tagging groups the words based on their logical semantic categories, and encodes simpler word category information in a sequence. Also, as in \cite{singh-etal-2018-twitter,sharma2016shallow}, Language Tag and Part of Speech Tag have previously been provided as manual handcrafted features for a range of downstream syntactic and semantic tasks. 

In addition to the above tasks, Language Model pretraining has shown significant performance gains as reported by \cite{howard2018universal}. It captures various aspects of language such as long range dependencies \cite{linzen2016assessing}, word categories, and sentiment \cite{radford2017learning}.

Conforming with the linguistic hierarchical information, we first train our model to predict the language labels for each character trigram as per its token. This is followed by further training the model to predict the Part of Speech tag for each of its character trigram as per its token. Subsequently, the model is trained on Language Modelling task, in process training the LSTM Layer 2 to build over the LSTM Layer 1 inputs to learn meaningful sequence representation. Lastly, the model is trained to predict the sentiment of the input text based on the LSTM Layer 2 representation.

\begin{table}[t]
\centering
\begin{tabular}{@{}ll@{}}
\toprule
\textbf{Hyperparameter}      & \textbf{Value} \\ \midrule
Embedding Dimension & 64    \\
LSTM Cell Dimension & 64    \\
Dropout             & 0.2   \\
Learning Rate       & 0.04  \\
Batch Size          & 4     \\ \bottomrule
\end{tabular}
\caption{Model Hyperparameters}
\label{table:params}
\end{table}

\begin{table}[t]
\centering
\begin{tabular}{@{}ll@{}}
\toprule
\textbf{Character Encoding}      & \textbf{Accuracy} \\ \midrule
Character Unigram & 62.57    \\
Convolution over Character Unigrams & 64.23    \\
Character Trigram & 67.83    \\
Byte Pair Encoding(BPE)             & 64.61   \\ \bottomrule
\end{tabular}
\caption{Character Encoding Experiments}
\label{table:chenc}
\end{table}

{

\begin{table*}[ht]
\centering
\begin{tabular}{|l|c|c|c|c|}
\hline
\multicolumn{1}{|r|}{Model} & \multicolumn{1}{l|}{Accuracy} & \multicolumn{1}{l|}{Precision} & \multicolumn{1}{l|}{Recall} & \multicolumn{1}{l|}{F1-score} \\ \hline
SVM(Unigrams)               & 61.7                          & 0.579                          & 0.551                       & 0.565                         \\ \hline
SVM(Unigrams+Bigrams)       & 64.1                          & 0.609                          & 0.537                       & 0.566                         \\ \hline
MNB(Unigrams)               & 64.5                          & 0.748                          & 0.485                       & 0.588                         \\ \hline
MNB(Unigrams+Bigrams)       & 66.1                          & 0.698                          & 0.540                       & 0.609                         \\ \hline
SentiWordNet                & 51.5                          & -                              & -                           & 0.252                         \\ \hline
Char-trigram based LSTM \cite{DBLP:madanEmsemble}    & 65.2                          & 0.610                          & 0.563                       & 0.586                         \\ \hline
Vowel-Consonant based       & 62.8                          & 0.652                          & 0.522                       & 0.580                         \\ \hline
Sub-word\cite{DBLP:AmeyaPrabhuJSV16}                    & 69.2                          & 0.684                          & 0.623                       & 0.652                         \\ \hline
Our Approach                & 72.51\%          & 0.712           & 0.645        & 0.677          \\ \hline
\end{tabular}
\caption{Model Comparison}
\end{table*}

}

\begin{table}[t]
\centering
\begin{tabular}{@{}ll@{}}
\toprule
\textbf{Training}      & \textbf{Accuracy} \\ \midrule
From Scratch (No Curriculum) & 70.19    \\
+ POS + LangId Pretraining & 68.83    \\
+ LM Pretraining & 72.51    \\
LM only Pretraining & 72.16    \\
No Gradual Unfreezing             &  70.74  \\
No Discriminative Finetuning             & 71.68   \\
\bottomrule
\end{tabular}
\caption{Curriculum and Finetuning Experiments}
\label{table:curr}
\end{table}

\subsection{Transfer Learning}

As noted in earlier efforts \cite{howard2018universal} towards finetuning pretrained models for NLP tasks, aggressive finetuning can cause catastrophic forgetting, thus causing the model to simply fit over the target task and forget any capabilities gained during the pretraining stage. On the other hand, too cautious finetuning can cause slow convergence and overfitting. To this end, we experiment with different strategies which can be broadly categorized as:

\paragraph{Discriminative Finetuning :} As also noted by \cite{yosinski2014transferable}, different layers capture different types of information, and thus need to be optimised differently. In the context of our model, the embedding layer captures the individual character trigram information, the LSTM layer 1 is trained towards capturing the token level information such as Part of Speech and Language Tag, and the final LSTM Layer 2 is trained to capture the overall textual representation to perform Language Modelling and Sentiment Analysis. With this purview, similar to \cite{howard2018universal}, we propose optimizing different layers in our model to different extents, and keep lower step sizes for the deeper pretrained layers while finetuning on a downstream task. We thus split the parameters as $\{\theta_1, ..., \theta_l\}$ , where $\theta_i$ corresponds to the parameters of layer i, and optimize them with separate learning rates $\{\eta_1, ...., \eta_l\}$ . Also, when finetuning a pretrained layer for a downstream task, we keep $\eta_i < \eta_j; \forall i<j$.

 Thus, while finetuning the POS + Lang Id pretrained model for Language Modeling, we propose to keep the learning rates for Embedding Layer and LSTM Layer 1 lower than the LSTM Layer 2 weights. Similarly, when finetuning the Language Model for Sentiment Analysis, we keep the learning rates of the deeper layers lower than that of the shallower ones.

\paragraph{Gradual Unfreezing:} Similar to \cite{howard2018universal}, rather than updating all the layers together for finetuning, we explore gradual ordered unfreezing of layers. Thus, initially we freeze all the layers. Then starting from the last layer, we train the model for a certain number of epochs before unfreezing the layer below it. Thus for Sentiment Analysis finetuning, for the first epoch, only $\theta_{sentiment}$ receives the gradient updates, after which we unfreeze the $\theta_{lstm2}$, and subsequently unfreeze the lower layers in a similar manner.

\section{Experiments}

The input to the LSTM stack is the sequence of character trigram dense representations, which we keep as 64 dimensional vectors. We also explore other token representations such as sequence of unigrams, convolution over unigrams \cite{DBLP:AmeyaPrabhuJSV16}, and Byte Pair Encoding (BPE) \cite{sennrich2015neural}. BPE is an unsupervised approach towards subword decomposition, and has shown improvements in MT systems and summarization. We train our model from scratch for Sentiment Analysis using the above mentioned character encodings, and report the results in Table \ref{table:chenc}.

Our LSTM stack consists of two layers of bidirectional LSTMs, with 64 hidden state dimensions. We add a dropout layer with the dropout rate set to 0.2 between the LSTM layers to prevent overfitting. We experiment with average pooling and max pooling concatenation over hidden states for semantic prediction, similar to \cite{howard2018universal}, and observe increase in model accuracy by 2.2\% on sentiment analysis.

To evaluate our baseline for curriculum training experiments, we initially train the model from scratch on the single target task (Sentiment Analysis) for 25 epochs. We approach the evaluation of our curriculum by training the model sequentially for four subtasks - Language Identification, POS Tagging, Language Modelling and Sentiment Analysis. We evaluate the strategy of pretraining with only POS Tagging and Language Identification, and observe similar performance as no curriculum training. We hypothesize the potential reasons for this drop and find a significant divergence in character trigram occurance between the Source Tasks (POS + Lang Id) and Target Task(Sentiment Analysis). This experiement highlights the importance of inclusion of language model pretraining for better token level representation learning, and provides a better model prior for sequence representation (LSTM Layer 2 output). We experiment with only Language Modelling as pretraining task, and observe significant gains over no curriculum strategy.

We note the convergence of our model with and without curriculum training, and observe that the curriculum training regimen causes faster convergence, as has been observed in previous works \cite{bengio2009curriculum,howard2018universal}. This is expected as the model is pretrained on prior tasks already have a general purpose representation learning, and only needs to adapt to the idiosyncrasies of the target task, i.e. Sentiment Analysis in this case.

As discussed in Section 4.3, for our transfer learning optimization experiments, we segment the optimization of different parameters of our model with different learning rates, in order to limit catastrophic forgetting and interference among the tasks, as proposed by \cite{howard2018universal}. We segregate our model parameters in the following 4 groups:
\begin{itemize}
\item Emb Layer
\item LSTM Layer 1
\item LSTM Layer 2
\item Sentiment Linear Map
\end{itemize}
We set the learning rate of the previous layer $\eta_{l-1}$ = $\eta_l / 2.0$. For our gradual unfreezing experiments, we unfreeze the lower layer after training the model for 1 epoch with the lower layer unfrozen. 


\section{Future Work}

In future work, we would like to explore word normalization and miltilingual embeddings in conjunction to learning representations from scratch. Another line of potential study could be investigation into why BPE is able to lead to performance gains as in monolingual domains, but fails in the codemixed multilingual tasks in our experiments. We would also like to explore convolution over character embeddings as a method to further circumvent the out of vocabulary problem with codemixed social media data.

We also plan to explore better representation learning for the semantic tasks. One particular direction we plan to explore  attention over the LSTM layer 2 as a weighted peek into the intermediate hidden states for the semantic classification task.

In future, we would like to experiment with more syntactic tasks like dependency label predictions, and also study more semantic tasks like aggression detection. Codemixed domains suffer from severe resource scarcity, and thus vocabulary divergence between various datasets proves as a roadblock to generalizable models, as observed in POS + Language Id pretraining experiments.

\bibliographystyle{named}
\bibliography{ijcai19}

\end{document}